\documentclass[letterpaper]{article}
\usepackage{aaai2027}
\usepackage[hyphens]{url}
\usepackage{graphicx}
\usepackage{natbib}
\usepackage{caption}
\usepackage{amsmath}
\usepackage{amssymb}
\usepackage{booktabs}
\usepackage{multirow}

\DeclareMathOperator{\sg}{sg}

\title{JSGS: JPEG State-Guided Supervision for 3D Gaussian Splatting from Mixed-Quality Views}

\author{\mdseries\small 
  Jinhua Cui$^1$ \quad Anhong Wang$^1$ \quad Kai Hu$^1$ \quad Donghan Bu$^1$ \quad 
  Peihao Li$^2$ \quad Tammam Tillo$^1$ \quad Hao Jing$^1$ \quad Shiao Xu$^1$
}
\affiliations{
  $^1$Taiyuan University of Science and Technology \\
  $^2$Chn Energy Digital Intelligence Technology Development (Beijing)
}

\begin{document}

\maketitle

\begin{abstract}
Standard 3D Gaussian Splatting (3DGS) assumes that every input image faithfully samples scene radiance.
However, mixed-quality JPEG images violate this assumption because compression-induced blocking and ringing artifacts can corrupt updates to Gaussians shared across views.
To address this problem, we propose JPEG State-Guided Supervision for 3D Gaussian Splatting from Mixed-Quality Views (JSGS).
JSGS uses luminance and chrominance quantization tables stored in each JPEG file to construct a view-specific JPEG observation operator.
This operator encodes and decodes each rendered view for domain-matched comparison with the corresponding decoded input image.
The luminance quantization table supplies continuous
weights within a fixed middle frequency band. A loss in
the low frequency band anchors coarse structure, while
the weighted middle frequency loss redistributes
supervision among the selected DCT coordinates. The
resulting block disagreement also guides the Gaussian
Controller to regularize small primitives with high
opacity in disagreement regions.
Across seven scenes and three mixed-quality schedules, JSGS achieves the lowest mean LPIPS and the highest mean SSIM under every schedule while rendering at approximately 150 FPS.Code:
\url{https://github.com/Jayden-Cui/JSGS}.
\end{abstract}

\section{Introduction}

\begin{figure*}[t]
\centering
\includegraphics[width=0.98\textwidth]{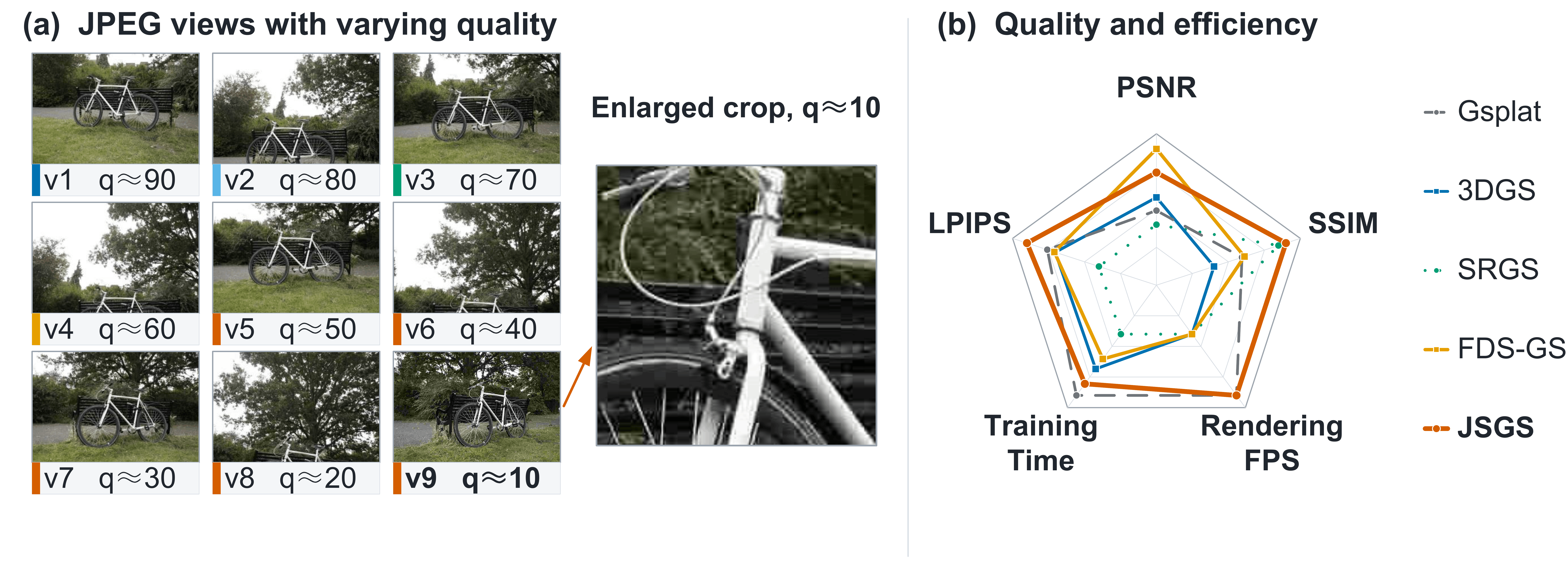}
\caption{
JPEG quality variation across views and aggregate
performance.
(a) Nine training views from the \textit{bicycle} scene
have different JPEG compression qualities, denoted by
$q$; lower $q$ indicates stronger compression. The crop
at $q \approx 10$ highlights blocking and ringing around
the bicycle frame and wheel.
(b) The radar chart summarizes PSNR, SSIM, and LPIPS
over all combinations of seven scenes and three schedules,
together with training time and rendering speed. JSGS
achieves the lowest aggregate LPIPS, the highest aggregate
SSIM, and ties for the highest reported rendering speed.
Raw values are reported
in Tables~\ref{tab:results} and~\ref{tab:timing}.
}
\label{fig:intro_teaser}
\end{figure*}

3D Gaussian Splatting (3DGS) represents a scene with explicit Gaussian primitives optimized using photometric supervision from multiple views \citep{kerbl2023gaussian}. This optimization treats each decoded training image as a faithful observation of scene radiance. In practical image collections, however, JPEG quality can vary from one view to another, producing blocking and ringing whose strength and spatial support differ across views. Because the training views jointly update a shared set of geometry, opacity, and appearance parameters, these structured residuals can alter primitives that also contribute to other views. The resulting challenge therefore involves both degradation within individual images and unequal supervision of a shared 3D representation across views. Figure~\ref{fig:intro_teaser} illustrates the variation in JPEG quality across training views and summarizes reconstruction quality and efficiency.

Recent work has increasingly relaxed the assumption that every 3DGS input provides a faithful observation. Degradation-specific methods model motion or defocus blur and challenging illumination, while more general approaches improve robustness to low resolution, noise, blur, and compression artifacts \citep{lee2024deblurring,zhao2024badgaussians,cui2025luminancegs,lin2025hqgs}. Anti-aliasing methods address a related but distinct mismatch caused by changes in rendering scale rather than by the stored training views \citep{yu2024mipsplatting}. When JPEG appears in degraded-scene 3DGS, however, it is typically treated as one member of a broad degradation set, rather than through the per-view quantization tables that specify the quantization steps at different block-DCT coordinates. Recent compression-aware 3DGS considers frame-wise quality variation in long videos \citep{song2026compsplat}, but does not use the JPEG quantization table of each image to condition supervision. Consequently, how per-view JPEG quantization information should condition the supervision of Gaussians shared across cameras remains underexplored.

To address this open question, we propose JSGS for 3D Gaussian Splatting from mixed-quality views. Our key insight is that the quantization tables stored in each JPEG file should influence both how a rendering is compared with its observation and how frequency disagreement updates the shared Gaussian representation. At the observation level, the JPEG Observation Operator (JO) passes each rendered image through a differentiable JPEG process using the corresponding luminance and chrominance tables before comparing it with the decoded input. At the frequency level, JSGS anchors coarse structure in the low frequency band and uses the luminance table to weight residuals in the middle frequency band. At the representation level, these weighted residuals are projected to visible Gaussians, where they regularize small primitives with high opacity in regions where disagreement remains. Across seven scenes and three schedules that mix JPEG quality levels, JSGS achieves the lowest LPIPS in all 21 combinations and the highest mean SSIM for each schedule.

In summary, our contributions are as follows:
\begin{itemize}
    \item We formulate mixed JPEG quality as unequal observation fidelity in 3DGS, where multiple views jointly update a shared Gaussian representation.
    \item We design a JPEG Observation Operator and complementary
    losses in the low and middle frequency bands. Continuous
    quantization-table weights redistribute residual contributions within
    the fixed middle frequency band, and the resulting block
    disagreement guides Gaussian regularization.
    \item We test seven scenes and three schedules. JSGS achieves the lowest LPIPS in all combinations and the highest mean SSIM under all three schedules.
\end{itemize}

\section{Related Work}

\subsection{3DGS under Degraded Observations}

Research on degraded 3DGS covers low resolution and sparse
views \citep{feng2024srgs,wan2025s2gaussian}, blur
\citep{lee2024deblurring,chen2024deblurgs,zhao2024badgaussians},
illumination and low light \citep{cui2025luminancegs,
ye2024gaussianindark,li2024chaos}, and transient distractors
\citep{kulhanek2024wildgaussians,sabour2025spotlesssplats}.
HQGS and ReSplat address several degradation types through
structural guidance or restoration assisted reconstruction
\citep{lin2025hqgs,yoon2026resplat}. HQGS includes compression
artifacts but does not use the quantization table stored in each
JPEG file. CompSplat uses frame quality in unposed videos to
regulate weighting, densification, and pruning
\citep{song2026compsplat}. JSGS instead uses each JPEG view's
quantization tables to supervise the shared representation
rather than assigning a scalar weight to each frame.

\subsection{Quantization Information in JPEG Images}

JPEG stores block DCT quantization tables in each file header
\citep{itu1992jpeg}. Prior restoration operates in the pixel
domain \citep{dong2015arcnn,zhang2017dncnn}, combines pixel
and DCT representations \citep{guo2016dual,zheng2020idcn},
or learns continuous cosine coefficients for decoding
\citep{han2024jdec}. QGAC conditions artifact correction on
the input quantization matrix, whereas FBCNN estimates a
quality factor when the compression setting is unknown
\citep{ehrlich2020qgac,jiang2021fbcnn}. These methods restore
images independently. JSGS instead uses the luminance and
chrominance tables to determine how each decoded observation
supervises a shared 3D representation without first restoring
the training image.

\subsection{Frequency and Density Control in 3DGS}

FreGS, Mip-Splatting, FDS-GS, and Revising Densification use
frequency, scale, or pixel error to improve representation and
density control \citep{zhang2024fregs,yu2024mipsplatting,
zeng2025fdsgs,rotabulo2024revising}. Other strategies use
gradient magnitude, coverage, propagation, or stochastic
relocation to allocate Gaussians \citep{ye2024absgs,
zhang2024pixelgs,fang2024minisplatting,cheng2024gaussianpro,
kheradmand2024mcmc,lu2024scaffoldgs}. JSGS instead treats DCT
coordinates as observation evidence whose precision varies
with each JPEG quantization table.

\section{Method}

\subsection{Problem Setup}

Let $\{(I_i,\pi_i,T_i^Y,T_i^C)\}_{i=1}^{N}$ denote the JPEG training data, where $I_i$ is the decoded image, $\pi_i=(K_i,R_i,t_i)$ denotes the calibrated camera parameters for view $i$, and $T_i^Y$ and $T_i^C$ are the luminance and chrominance quantization tables parsed from the corresponding JPEG file. Here $K_i$ is the intrinsic matrix, while $R_i$ and $t_i$ are the extrinsic rotation and translation. Following standard 3DGS \citep{kerbl2023gaussian}, the shared scene representation is $\Theta=\{(\mu_j,q_j,s_j,\alpha_j,c_j)\}_{j=1}^{M}$, and rendering it under $\pi_i$ produces $\hat I_i$. Here $\mu_j$, $q_j$, $s_j$, $\alpha_j$, and $c_j$ denote the center, rotation, scale, opacity, and appearance of Gaussian $j$, respectively, while $\mathcal R(q_j)$ denotes its rotation matrix.

At pixel $p$, let $a_{ij}(p)$ and $t_{ij}(p)$ denote the projected footprint and accumulated transmittance of Gaussian $j$. Its compositing weight is therefore $t_{ij}(p)\alpha_j a_{ij}(p)$. All training views optimize the same $\Theta$; JSGS uses the quantization tables associated with view $i$ to determine how that observation supervises the shared representation. Standard rasterization and table parsing conventions are provided in the supplementary material.

\begin{figure*}[t]
\centering
\includegraphics[width=0.98\textwidth]{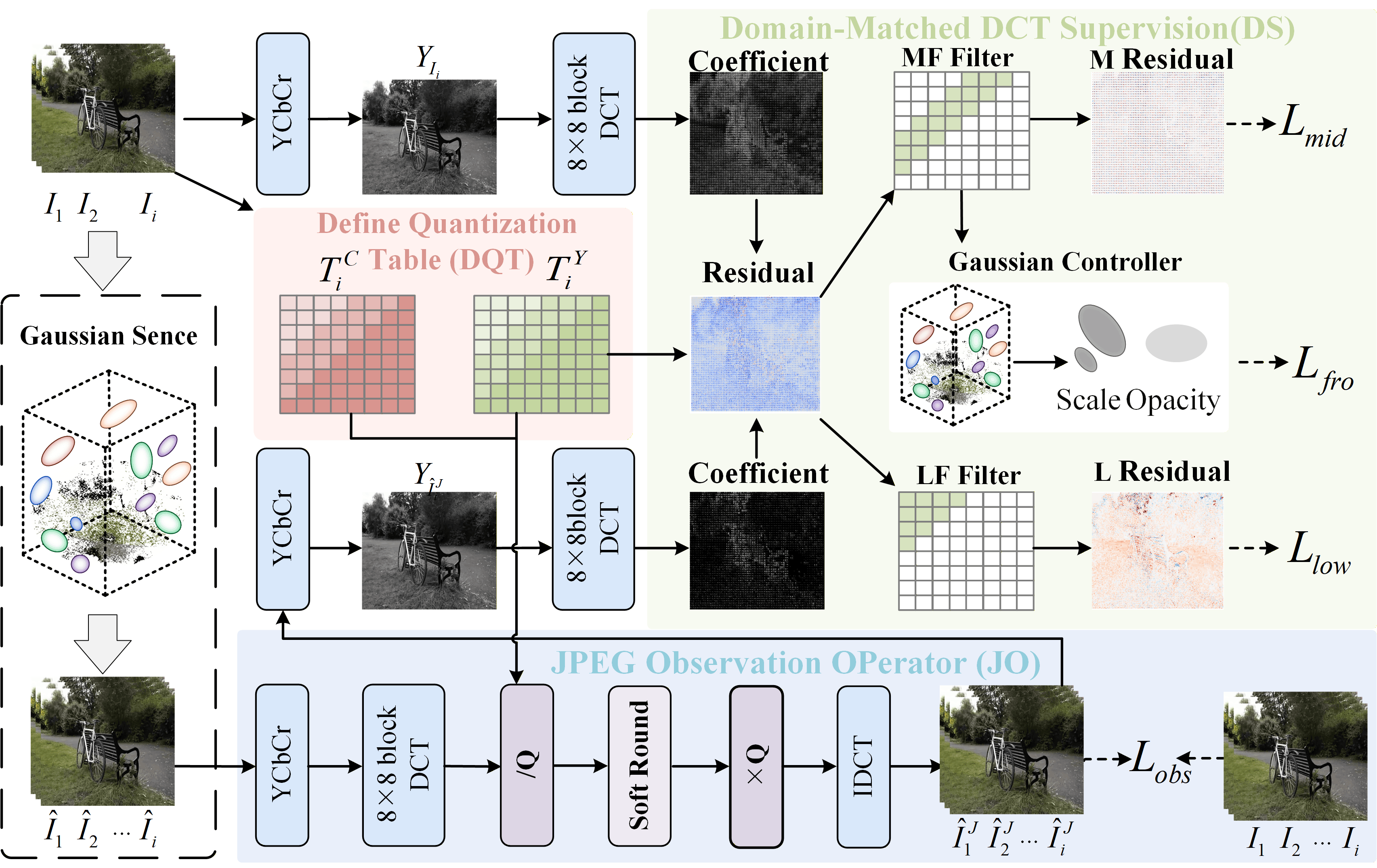}
\caption{Overview of JSGS. $I_i$, $\hat I_i$, and $\hat I_i^J$ denote the input JPEG images, rendered images, and corresponding JPEG Observation Operator (JO) outputs, respectively. The Define Quantization Table (DQT) block reads $T_i^Y$ and $T_i^C$ from each JPEG header. JO applies differentiable JPEG encoding and decoding operations to $\hat I_i$ using these tables, producing $\hat I_i^J$; $L_{\mathrm{obs}}$ compares $\hat I_i^J$ with $I_i$. Domain-Matched DCT Supervision (DS) forms the luminance DCT residual between $I_i$ and $\hat I_i^J$ and computes $L_{\mathrm{low}}$ and $L_{\mathrm{mid}}$. DS also projects MF disagreement to the Gaussian Controller (GC) for $L_{\mathrm{fro}}$.}
\label{fig:pipeline}
\end{figure*}

\subsection{Framework Overview}

Figure~\ref{fig:pipeline} summarizes the design of JSGS.
JSGS treats the quantization tables of each view as JPEG
state rather than a scalar quality label. JO applies this state
to the rendered image $\hat I_i$, producing the JPEG
observation $\hat I_i^J$ for image-domain comparison with
the decoded input $I_i$. DS applies frequency-domain
supervision to the same image pair, using the luminance table
to weight DCT residuals and identify disagreement for GC.
Thus, both image-domain supervision and Gaussian
regularization depend on the compression state of the current view.

\subsection{JPEG Compression State and Frequency Reliability}

The luminance and chrominance quantization tables recorded
in each file header jointly define the JPEG compression state
of view $i$. JO uses both tables. The LF and MF losses
operate on luminance DCT residuals. Only MF weighting and GC use reliability derived from $T_i^Y$.
Implementation and protocol details are provided in
Supplementary Section~C.

Let $(u,v)\in\mathcal K:=\{0,\ldots,7\}^2$ index a DCT table in natural coordinate order, and let $F(u,v)=u+v$. For the 8-bit tables considered here, $T_i^Y(u,v)\in\{1,\ldots,255\}$ and $F(u,v)\in\{0,\ldots,14\}$. We convert the luminance quantization step into a monotone reliability weight:
\begin{equation}
\omega_i(u,v)=
\exp\!\left[
-2\frac{T_i^Y(u,v)-1}{254}
-\frac{F(u,v)}{14}
\right].
\label{eq:omega}
\end{equation}
Smaller quantization steps produce larger weights, while the second factor discounts higher DCT coordinates. The constants 254 and 14 normalize the two finite ranges, and the coefficient 2 controls DQT-dependent decay. Because subsequent weighted means normalize $\omega_i$, it redistributes supervision within the selected band according to relative DQT reliability. The MF support remains fixed across views, while the recorded table values determine the influence of each selected coordinate. We use $\epsilon_{\mathrm{sm}}$ to control the smooth absolute penalty near zero. The constants $\epsilon_{\mathrm{DCT}}^{\mathrm{loss}}$, $\epsilon_{\mathrm{DCT}}^{\mathrm{block}}$, and $\epsilon_{\mathrm{proj}}$ are denominator safeguards for the DCT losses, the block score, and Gaussian attribution, respectively. Their values are reported in the experimental setup and Supplementary Section~C.

\subsection{JPEG Observation Operator}

For view $i$, JO maps the rendered image through the
quantization tables recorded in the corresponding JPEG file:
\begin{equation}
\hat I_i^J=C_{\mathrm{JPEG}}(\hat I_i;T_i^Y,T_i^C).
\label{eq:jpeg_forward}
\end{equation}
The output $\hat I_i^J$ is the JPEG observation used by
$L_{\mathrm{obs}}^i$.

JO applies a fixed $4{:}2{:}0$ JPEG transform using the
recorded quantization tables. Let $\mathcal T_{\mathrm{enc}}$ and
$\mathcal T_{\mathrm{dec}}$ collect the standard fixed JPEG
operations outside DCT and quantization, and let
$\mathcal{D}$ denote the orthonormal $8\times8$ block
DCT. The tuple
$\mathbf T_i=(T_i^Y,T_i^C,T_i^C)$ assigns the recorded tables
to the luminance and chrominance components.

For a scalar $z\in\mathbb{R}$, where $z$ denotes one
DCT coefficient divided by its corresponding quantization
step before rounding, we define the elementwise soft
rounding function:
\begin{equation}
r_{\mathrm{soft}}(z)
=
\operatorname{round}(z)
+
\left[z-\operatorname{round}(z)\right]^3.
\label{eq:soft_round}
\end{equation}
The backward pass treats $\operatorname{round}$ as zero-gradient. The JO is defined as follows:
\begin{equation}
\begin{aligned}
u_i &= \mathcal D\!\left(\mathcal T_{\mathrm{enc}}(\hat I_i)\right),
\qquad u_i=(u_i^Y,u_i^{\mathrm{Cb}},u_i^{\mathrm{Cr}}),\\
\hat u_i &= \operatorname{Quant}_i(u_i)
=\mathbf T_i\,r_{\mathrm{soft}}\!\left(\frac{u_i}{\mathbf T_i}\right),\\
\hat I_i^J &= \operatorname{clip}\!\left(
\mathcal T_{\mathrm{dec}}\!\left(\mathcal D^{-1}(\hat u_i)\right),0,1\right).
\end{aligned}
\label{eq:jpeg_operator}
\end{equation}
All coefficient operations are elementwise, and the
reconstructed RGB output is clipped to $[0,1]$. Exact
color conversion, chroma sampling, block assembly, inverse
transform, clipping, and backward conventions are provided
in Supplementary Section~A.

For two RGB images
$X,Y\in[0,1]^{H\times W\times3}$, we write the standard
$\ell_1$--D-SSIM photometric loss as
\begin{equation}
\begin{aligned}
\ell_{\mathrm{ph}}(X,Y)
&=
(1-\lambda_{\mathrm{ssim}})
\frac{\|X-Y\|_1}{3HW}
\\
&\quad+
\lambda_{\mathrm{ssim}}
\left[
1-\operatorname{SSIM}(X,Y)
\right].
\end{aligned}
\label{eq:photometric}
\end{equation}
Here $\lambda_{\mathrm{ssim}}\in[0,1]$ controls the mixture. The JPEG observation loss is
\begin{equation}
L_{\mathrm{obs}}^i=\ell_{\mathrm{ph}}(\hat I_i^J,I_i).
\label{eq:obs_loss}
\end{equation}
This loss compares the decoded input with the rendered image
after JO processing.

\subsection{Domain-Matched DCT Supervision}

Matched observation processing places
$\widehat I_i^J$ and $I_i$ in the same decoded JPEG
domain, but it does not distinguish DCT coordinates with
different quantization steps. JSGS therefore uses the
recorded luminance quantization table $T_i^Y$ to describe
the JPEG state of view $i$ at each DCT coordinate. Together
with the coordinate index, this table defines the continuous
weights $\omega_i(u,v)$ within the fixed middle frequency
band, while the low frequency branch uses a fixed mask. We
compute the DCT residuals between $I_i$ and
$\widehat I_i^J$.

For an RGB image
$X\in[0,1]^{H\times W\times3}$, we define the 8-bit
luminance map used for DCT supervision as
\begin{equation}
\mathcal{Y}_{8}(X)
=
255\left(
0.299X_R+0.587X_G+0.114X_B
\right).
\label{eq:jsgs_luminance}
\end{equation}
Before applying the block DCT, we perform the 8-bit level
shift as follows:
\begin{equation}
Y_{I_i}
=
\mathcal{Y}_{8}(I_i)-128,
\qquad
Y_{\widehat I_i^J}
=
\mathcal{Y}_{8}(\widehat I_i^J)-128.
\label{eq:jsgs_level_shift}
\end{equation}
Let $\mathcal{B}_i$ denote the set of valid,
nonoverlapping $8\times8$ blocks, and let
$\mathcal{P}_b$ extract the luminance block indexed by
$b$. Using the orthonormal block DCT $\mathcal{D}$, we
define:
\begin{equation}
\begin{aligned}
C_I^{i,b}(u,v)
&=
\left[
\mathcal{D}
\left(
\mathcal{P}_b(Y_{I_i})
\right)
\right]_{u,v},
\\
C_J^{i,b}(u,v)
&=
\left[
\mathcal{D}
\left(
\mathcal{P}_b(Y_{\widehat I_i^J})
\right)
\right]_{u,v}.
\end{aligned}
\label{eq:jsgs_dct_coefficients}
\end{equation}
Here $C_I^{i,b}(u,v)$ and $C_J^{i,b}(u,v)$ are the
luminance DCT coefficients of the decoded observation and
the JPEG Observation Operator output, respectively.
Unless stated otherwise, sums over $b,u,v$ range over
$\mathcal{B}_i\times\mathcal{K}$.

We compute scaled DCT coefficient residuals and apply a
smooth absolute penalty:
\begin{equation}
\begin{aligned}
\Delta_i^b(u,v)
&=
\frac{
C_J^{i,b}(u,v)-C_I^{i,b}(u,v)
}{255},
\\
\psi_{\epsilon_{\mathrm{sm}}}(x)
&=
\sqrt{x^2+\epsilon_{\mathrm{sm}}^2}
-\epsilon_{\mathrm{sm}},
\qquad
\epsilon_{\mathrm{sm}}>0.
\end{aligned}
\label{eq:jsgs_dct_residual}
\end{equation}
Here $\Delta_i^b(u,v)$ is the signed DCT coefficient
discrepancy after fixed scaling, and
$\psi_{\epsilon_{\mathrm{sm}}}$ is a smooth approximation
to the absolute value.
Division by $255$ sets a fixed coefficient scale, while
$\Delta_i^b(u,v)$ remains an unconstrained signed residual.

The low and middle frequency masks and the corresponding
view dependent middle frequency weights are
\begin{equation}
\begin{aligned}
M_{\mathrm{low}}(u,v)
&=
\mathbf{1}\{F(u,v)\leq3\},
\\
M_{\mathrm{mid}}(u,v)
&=
\mathbf{1}\{3<F(u,v)\leq5\},
\\
W_{\mathrm{mid},i}(u,v)
&=
M_{\mathrm{mid}}(u,v)\omega_i(u,v).
\end{aligned}
\label{eq:jsgs_frequency_masks}
\end{equation}
Here $\mathbf{1}\{\cdot\}$ equals one when its condition
holds and zero otherwise. Because $F(u,v)=u+v$ is integer
valued, $M_{\mathrm{mid}}$ selects exactly the coordinates
with $F(u,v)\in\{4,5\}$. Both frequency masks are fixed
across views. All selected MF coordinates remain active,
and the positive continuous weights $\omega_i(u,v)$ adapt
their relative contributions to view $i$.

For each block $b$, JSGS defines the weighted middle
frequency error:
\begin{equation}
e_i^b
=
\sum_{u,v}
W_{\mathrm{mid},i}(u,v)
\psi_{\epsilon_{\mathrm{sm}}}
\left(
\Delta_i^b(u,v)
\right).
\label{eq:jsgs_mid_block_error}
\end{equation}
JSGS then uses the following two DCT losses for view
$i$:
\begin{equation}
\begin{aligned}
L_{\mathrm{low}}^i
&=
\frac{
\sum_{b,u,v}
M_{\mathrm{low}}(u,v)
\psi_{\epsilon_{\mathrm{sm}}}
\left(\Delta_i^b(u,v)\right)
}{
\sum_{b,u,v}
M_{\mathrm{low}}(u,v)
+
\epsilon_{\mathrm{DCT}}^{\mathrm{loss}}
},
\\
L_{\mathrm{mid}}^i
&=
\frac{
\sum_{b\in\mathcal{B}_i}e_i^b
}{
\sum_{b,u,v}
W_{\mathrm{mid},i}(u,v)
+
\epsilon_{\mathrm{DCT}}^{\mathrm{loss}}
}.
\end{aligned}
\label{eq:jsgs_dct_losses}
\end{equation}
$L_{\mathrm{low}}^i$ anchors coarse luminance structure.
$L_{\mathrm{mid}}^i$ is a normalized weighted mean over
the fixed middle frequency band. Within this band,
$W_{\mathrm{mid},i}(u,v)$ scales the relative contribution
of each residual according to $\omega_i(u,v)$. Coordinates
with smaller quantization steps receive larger relative
weights. The normalization keeps the loss scale comparable
while allowing each DQT to determine which MF residuals
dominate supervision.

\begin{table*}[t]
\centering
\small
\begin{tabular*}{\textwidth}{@{\extracolsep{\fill}}lccccccccc@{}}
\toprule
\multirow{2}{*}{Method} & \multicolumn{3}{c}{Uniform} & \multicolumn{3}{c}{Gaussian} & \multicolumn{3}{c}{Descending}\\
\cmidrule(lr){2-4}\cmidrule(lr){5-7}\cmidrule(lr){8-10}
 & PSNR$\uparrow$ & SSIM$\uparrow$ & LPIPS$\downarrow$ & PSNR$\uparrow$ & SSIM$\uparrow$ & LPIPS$\downarrow$ & PSNR$\uparrow$ & SSIM$\uparrow$ & LPIPS$\downarrow$\\
\midrule
Gsplat & 25.40 & 0.812 & 0.200 & 25.53 & 0.812 & 0.197 & 25.61 & 0.816 & 0.186\\
3DGS & 25.72 & 0.798 & 0.216 & 25.69 & 0.788 & 0.218 & 25.82 & 0.802 & 0.200\\
SRGS & 25.25 & 0.835 & 0.309 & 25.20 & 0.834 & 0.308 & 25.35 & 0.838 & 0.299\\
FDS-GS & \textbf{26.55} & 0.811 & 0.214 & \textbf{26.57} & 0.814 & 0.211 & \textbf{26.70} & 0.820 & 0.203\\
JSGS (ours) & 26.13 & \textbf{0.836} & \textbf{0.159} & 26.13 & \textbf{0.839} & \textbf{0.158} & 26.30 & \textbf{0.847} & \textbf{0.135}\\
\bottomrule
\end{tabular*}
\caption{Metrics are computed using uncompressed reference images and averaged over seven scenes for each mixed-quality schedule. JSGS achieves the lowest LPIPS in all 21 combinations of scenes and schedules. It also achieves the highest mean SSIM for every schedule. FDS-GS achieves the highest mean PSNR.}
\label{tab:results}
\end{table*}

\begin{figure*}[!t]
\centering
\includegraphics[width=\textwidth]{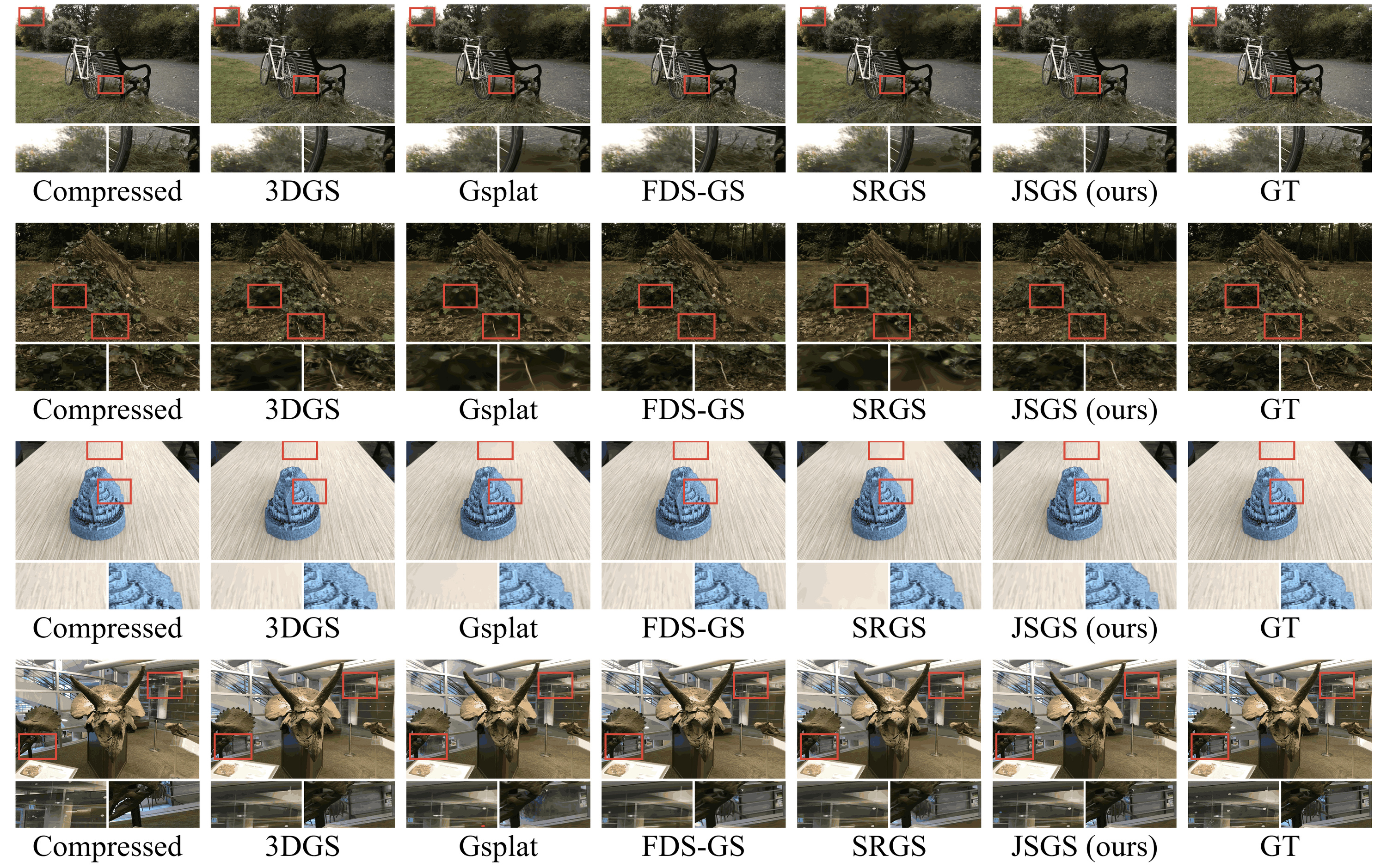}
\caption{Qualitative comparison on \textit{bicycle} and \textit{stump} from Mip-NeRF 360, and \textit{trex} and \textit{horns} from LLFF. The first column shows a JPEG-compressed input, the next five columns show renderings from different methods, and the final column shows the uncompressed ground truth (GT). The two small images below each main image show the enlarged crops.}
\label{fig:qualitative}
\end{figure*}

\subsection{Residual Projection and Gaussian Controller}

The weighted middle frequency error also supports
Gaussian attribution. For each block, JSGS defines:
\begin{equation}
r_i^b
=
\frac{
e_i^b
}{
\sum_{u,v}
W_{\mathrm{mid},i}(u,v)
+
\epsilon_{\mathrm{DCT}}^{\mathrm{block}}
}.
\label{eq:jsgs_block_score}
\end{equation}
$L_{\mathrm{mid}}^i$ aggregates $e_i^b$ over all valid
blocks, whereas $r_i^b$ preserves the location of block
$b$ for Gaussian attribution.
The score $r_i^b$ measures block-level DQT weighted
disagreement in the fixed middle frequency band. Assigning
this score to the pixels in block $b$ preserves the native
JPEG block support for Gaussian attribution.
Let $b_i(p)\in\mathcal B_i$ denote the non-overlapping block containing pixel $p$. We set $r_i(p)=r_i^{b_i(p)}$. Let $\Omega_i$ denote the pixel grid. All sums over $p$ below use $p\in\Omega_i$. JSGS defines visibility and pools residuals as follows:
\begin{equation}
\begin{aligned}
\nu_{ij}&=\sum_p t_{ij}(p)\alpha_j a_{ij}(p),
&V_i&=\{j:\nu_{ij}>0\},\\
\kappa_{ij}(p)&=t_{ij}(p)a_{ij}(p),\\
\bar r_{ij}&=\sg\!\left[
\frac{\sum_p\kappa_{ij}(p)r_i(p)}
{\sum_p\kappa_{ij}(p)+\epsilon_{\mathrm{proj}}}
\right].
\end{aligned}
\label{eq:gaussian_risk}
\end{equation}
Here $\nu_{ij}$ is the rendered contribution mass,
$V_i$ is the contributor set from the current rasterizer
pass, $\kappa_{ij}(p)$ is the footprint weight excluding
the current Gaussian's opacity $\alpha_j$, and
$\operatorname{sg}[\cdot]$ denotes stop
gradient. The pooled score $\bar r_{ij}$ is detached and
recomputed at each forward pass. Visibility, normalization,
and boundary conventions are detailed in Supplementary
Section~B.

Let $J_{ij}$ be the camera-to-pixel projection Jacobian
and $W_i$ the linear part of the world-to-camera map. We
define:
\begin{equation}
\begin{aligned}
A_{ij}
&=
\sg
\left[
J_{ij}W_i\mathcal R(q_j)
\right],
\\
\bar s_{ij}
&=
\sqrt{
\frac{1}{2}
\operatorname{tr}
\left(
A_{ij}
\operatorname{diag}(s_j)^2
A_{ij}^{\top}
\right)
}.
\end{aligned}
\label{eq:controller_scale}
\end{equation}
The quantity $\bar s_{ij}$ is the projected RMS scale
before fixed screen-space dilation. The stop gradient in
$A_{ij}$ freezes center, rotation, and camera variables.
The standard covariance-projection derivation and its
Frobenius-norm equivalence are provided in Supplementary
Section~B. We set $\tau_s=1.0$ pixel at the training
resolution. We compute:
\begin{equation}
L_{\mathrm{fro}}^i=
\frac{1}{|V_i|}
\sum_{j\in V_i}
\bar r_{ij}\alpha_j^2
\exp\!\left(-\frac{\bar s_{ij}}{\tau_s}\right),
\qquad |V_i|>0,
\label{eq:fro}
\end{equation}
With $\bar r_{ij}$ detached,
$L_{\mathrm{fro}}^i$ directly regularizes opacity and
projected scale where middle frequency disagreement
remains. Gradient directions, empty set handling, and
interactions with density control are provided in
Supplementary Section~B.

\subsection{Overall Objective}

The final training objective for view $i$ is
\begin{equation}
\begin{aligned}
L^i={}&\lambda_{\mathrm{base}}\ell_{\mathrm{ph}}(\hat I_i,I_i)
+\lambda_{\mathrm{obs}}L_{\mathrm{obs}}^i\\
&+\lambda_{\mathrm{low}}L_{\mathrm{low}}^i
+\lambda_{\mathrm{mid}}L_{\mathrm{mid}}^i
+\lambda_{\mathrm{fro}}L_{\mathrm{fro}}^i .
\end{aligned}
\label{eq:overall}
\end{equation}
During training, uniform stochastic view sampling estimates
$\frac{1}{N}\sum_{i=1}^{N}L^i$. The same fixed loss
weights are used for all scenes and quality schedules. These
terms supervise the JPEG observation, low frequency
structure, middle frequency detail weighted by the DQT, and
Gaussian regularization.

\section{Experiments}

\subsection{Experimental Setup}

\paragraph{Datasets.}
We evaluate four LLFF scenes \citep{mildenhall2019llff}---flower, fortress, horns, and trex---and three Mip-NeRF 360 scenes \citep{barron2022mipnerf360}---bicycle, garden, and stump. LLFF covers forward-facing captures, while Mip-NeRF 360 covers unbounded scenes with wider camera motion. For every scene, we reserve one uncompressed view from each group of eight for testing. This gives a 7:1 train-test split. We compress only the training views.

\paragraph{Implementation details.}
All methods are trained for 10,000 iterations. We evaluate the models at iterations 3,000, 7,000, and 10,000 and report the results at iteration 10,000 in the main tables. JSGS uses the same Gsplat backbone, initialization, optimizer settings, and density control configuration as the Gsplat baseline. We use $\lambda_{\mathrm{ssim}}=0.2$, $\lambda_{\mathrm{obs}}=0.2$, $\lambda_{\mathrm{low}}=0.02$, $\lambda_{\mathrm{mid}}=0.03$, and $\lambda_{\mathrm{fro}}=0.05$. We set $\epsilon_{\mathrm{sm}}=10^{-3}$ and $\tau_s=1.0$ pixel at the training resolution. To avoid division by zero, we add $10^{-6}$ to the denominators of the DCT losses, block score, and Gaussian attribution, corresponding to $\epsilon_{\mathrm{DCT}}^{\mathrm{loss}}$, $\epsilon_{\mathrm{DCT}}^{\mathrm{block}}$, and $\epsilon_{\mathrm{proj}}$, respectively. The same JSGS parameter set is used for every scene and quality schedule. Each run uses one NVIDIA GeForce RTX 5090 GPU with 32 GB memory. Supplementary Section~C provides more information regarding the training process, density control, JPEG generation, and evaluation configuration.

The mixed-quality regime draws from qualities 10--90 in steps of 10. Uniform assigns equal mass to all nine qualities. Gaussian weights nine positions in $[-2,2]$ by $\exp(-x^2/2)$. Descending uses weights $9{:}8{:}\cdots{:}1$ from $q=90$ to $q=10$. Fixed-quality controls set every training view to $q\in\{30,50\}$. Mixed schedules combine several JPEG states within one reconstruction, whereas fixed-quality controls remove this variation. Supplementary Section~C consolidates these implementation details.

\paragraph{Baselines and metrics.}
We compare Gsplat \citep{ye2025gsplat}, standard 3DGS, SRGS \citep{feng2024srgs}, FDS-GS \citep{zeng2025fdsgs}, and JSGS under both regimes. We compute PSNR, SSIM, and LPIPS using uncompressed reference images \citep{wang2004ssim,zhang2018lpips}. PSNR measures pixel fidelity, SSIM measures structural similarity, and LPIPS measures perceptual distance. We construct inputs using scalar JPEG encoder settings, while JSGS conditions training on the file-header quantization tables rather than the scalar quality labels.

\subsection{Comparison with Existing Methods}

\paragraph{Quantitative comparison.}
The mixed-quality comparison evaluates reconstruction under three schedules spanning nine JPEG quality levels. Motivated by prior work on frame-wise variations in compression quality~\citep{song2026compsplat}, the schedules assign different proportions of the training views to these quality levels, enabling comparison across distinct JPEG-quality distributions.

As shown in Table~\ref{tab:results}, relative to Gsplat, JSGS lowers average LPIPS by $20.7\%$, $20.1\%$, and $27.1\%$ across the three schedules. It also achieves the highest schedule-average SSIM. FDS-GS attains $0.40$--$0.44$ dB higher PSNR than JSGS, whereas JSGS achieves lower LPIPS and higher SSIM in every schedule. These metric-specific differences remain consistent as the distribution of JPEG qualities changes.

\paragraph{Qualitative comparison.}
Figure~\ref{fig:qualitative} compares four scenes using aligned image regions across all methods. The bicycle and horns examples emphasize thin contours, while stump and trex expose repeated textures and occlusion boundaries. The two crops below each result make blocking and ringing differences directly comparable. The aggregate results complement these local observations.

\subsection{Ablation Study}

Table~\ref{tab:ablation} compares five configurations of the three JSGS components: JO, DS, and GC.

\begin{table}[ht]
\centering
\small
\setlength{\tabcolsep}{4pt}
\begin{tabular*}{\columnwidth}{@{\extracolsep{\fill}}lcccccc@{}}
\toprule
Variant & JO & DS & GC & PSNR$\uparrow$ & SSIM$\uparrow$ & LPIPS$\downarrow$\\
\midrule
Baseline & -- & -- & -- & 26.02 & 0.849 & 0.118\\
JO only & $\checkmark$ & -- & -- & 26.09 & 0.860 & 0.105\\
JO + DS & $\checkmark$ & $\checkmark$ & -- & 26.32 & 0.860 & 0.101\\
JO + GC & $\checkmark$ & -- & $\checkmark$ & 26.13 & \textbf{0.866} & 0.104\\
Full JSGS & $\checkmark$ & $\checkmark$ & $\checkmark$ & \textbf{26.38} & \textbf{0.866} & \textbf{0.099}\\
\bottomrule
\end{tabular*}
\caption{Ablation on LLFF flower under the Gaussian mixed-quality schedule. JO, DS, and GC denote the JPEG Observation Operator, Domain-Matched DCT Supervision, and Gaussian Controller, respectively.}
\label{tab:ablation}
\end{table}

As shown in Table~\ref{tab:ablation}, adding JO lowers LPIPS from $0.118$ to $0.105$, showing the benefit of matching the rendered observation to the JPEG state. Relative to the JO-only variant, adding DS raises PSNR by $0.23$ dB and lowers LPIPS by $0.004$, while adding GC raises SSIM by $0.006$. Combining all three components achieves the best PSNR and LPIPS and ties the best SSIM. These gains are consistent with the complementary roles of DCT supervision and Gaussian regularization.

Both DS and GC operate on the JO output; accordingly, every configuration containing either component also includes JO. Because DS combines $L_{\mathrm{low}}$ and $L_{\mathrm{mid}}$, its column evaluates their joint effect rather than the two frequency bands separately.

\subsection{Robustness and Efficiency}

\paragraph{Fixed-quality controls.}
Table~\ref{tab:fixed_quality} removes cross-view quality variation and reports the $q=30$ and $q=50$ controls. JSGS achieves the lowest LPIPS at both qualities. It reduces LPIPS relative to Gsplat by $13.7\%$ and $15.4\%$, respectively. FDS-GS achieves the highest PSNR, while SRGS achieves the highest SSIM at both qualities. The LPIPS advantage therefore persists when all training views share one JPEG quality.

\begin{table}[h]
\centering
\small
\setlength{\tabcolsep}{1.5pt}
\begin{tabular}{lcccccc}
\toprule
\multirow{2}{*}{Method} & \multicolumn{3}{c}{$q=30$} & \multicolumn{3}{c}{$q=50$}\\
\cmidrule(lr){2-4}\cmidrule(lr){5-7}
 & PSNR$\uparrow$ & SSIM$\uparrow$ & LPIPS$\downarrow$ & PSNR$\uparrow$ & SSIM$\uparrow$ & LPIPS$\downarrow$\\
\midrule
Gsplat & 24.94 & 0.767 & 0.232 & 25.09 & 0.778 & 0.213\\
3DGS & 25.46 & 0.780 & 0.235 & 25.76 & 0.795 & 0.214\\
SRGS & 25.24 & \textbf{0.830} & 0.322 & 25.34 & \textbf{0.836} & 0.307\\
FDS-GS & \textbf{25.86} & 0.795 & 0.234 & \textbf{26.04} & 0.807 & 0.215\\
JSGS (ours) & 25.48 & 0.791 & \textbf{0.200} & 25.73 & 0.806 & \textbf{0.180}\\
\bottomrule
\end{tabular}
\caption{Seven-scene fixed-quality means at $q=30$ and $q=50$.}
\label{tab:fixed_quality}
\end{table}

\paragraph{Mixed-quality interpretation.}
JSGS achieves the lowest LPIPS in all $21$ mixed-quality and all $14$ fixed-quality scene-level comparisons. Under mixed quality, its average LPIPS reduction ranges from $20.1\%$ to $27.1\%$. At $q=30$ and $q=50$, the reduction ranges from $13.7\%$ to $15.4\%$. Under fixed quality, all views introduce a common compression level. Mixed schedules instead produce view-dependent residuals whose updates meet in shared Gaussians. A low-quality view can therefore alter primitives that higher-quality cameras also use. JSGS conditions these updates through observation matching, frequency masks, and the Gaussian Controller. The larger mixed-quality margin is consistent with the target setting in which several JPEG states jointly supervise one representation.

\begin{table}[h]
\centering
\small
\setlength{\tabcolsep}{4.5pt}
\begin{tabular}{@{}lcc@{}}
\toprule
Method & Training Time (min)$\downarrow$ & Rendering Speed (FPS)$\uparrow$\\
\midrule
Gsplat      & $\mathbf{\approx 3}$  & $\mathbf{\approx 150}$\\
3DGS        & $\approx 19$          & $\approx 80$\\
SRGS        & $\approx 40$          & $\approx 80$\\
FDS-GS      & $\approx 25$          & $\approx 80$\\
JSGS (ours) & $\approx 10$          & $\mathbf{\approx 150}$\\
\bottomrule
\end{tabular}
\caption{Approximate training time and rendering speed.}
\label{tab:timing}
\end{table}
\FloatBarrier

\paragraph{Efficiency.}
Table~\ref{tab:timing} reports approximate training time and rendering speed. JSGS trains in approximately 10 minutes, which is slower than Gsplat but faster than 3DGS, SRGS, and FDS-GS. Its rendering speed is approximately 150 FPS and matches Gsplat.

\section{Conclusion and Limitations}

We study 3D Gaussian Splatting from mixed-quality JPEG views. JSGS uses per-view quantization tables to align observations, reweight DCT residuals, and regularize shared Gaussians. Across seven scenes and three schedules, it achieves the lowest LPIPS in every scene--schedule comparison and the highest mean SSIM for each schedule, while FDS-GS retains higher mean PSNR. The LPIPS advantage persists in both fixed-quality controls, and ablations support JO, DS, and GC. JSGS retains Gsplat's approximately 150-FPS rendering speed.

We have not considered double JPEG compression: the final JPEG header records only the quantization tables of the last encoding stage and cannot fully represent the preceding compression history. In addition, our data are produced by controlled JPEG encoding of existing scene datasets and do not yet cover data generated by diverse real-world communication devices, encoders, and transmission links.

\bibliography{references}

\end{document}